\documentclass[conference]{IEEEtran}
\usepackage{cite}
\usepackage{amsmath,amssymb,amsfonts}
\usepackage{algorithmic}
\usepackage{graphicx}
\usepackage{textcomp}
\usepackage{xcolor}
\usepackage{caption}

\graphicspath{{images/}}

\title{Luggage re-identification: A study of strategies towards a real-world application}
\begin{document}

%\author{\IEEEauthorblockN{Quentin Dhôte}
%\IEEEauthorblockA{\textit{CEA, List, Université Paris-Saclay} \\
%91120 Palaiseau, France \\
%quentin.dhote@cea.fr}
%\and
%\IEEEauthorblockN{Romaric Audigier}
%\IEEEauthorblockA{\textit{CEA, List, Université %Paris-Saclay} \\
%91120 Palaiseau, France \\
%romaric.audigier@cea.fr}
%}

\author{\IEEEauthorblockN{Quentin Dhôte, Romaric Audigier}
\IEEEauthorblockA{\textit{
Université Paris-Saclay, CEA, List, F-91120 Palaiseau, France
} \\
quentin.dhote@cea.fr, romaric.audigier@cea.fr}
}

\maketitle

\begin{abstract}
Luggage re-identification (ReID) is a critical task in video surveillance for forensics, security or border management (suspicious or abandoned luggage recognition), as well as luggage management (lost luggage recovery). It enables the association of observations of the same luggage across disjoint camera views in complex environments such as airports, train stations and border crossing. While existing approaches primarily focus on appearance-based cropped image matching under relatively clean conditions, real-world scenarios are characterized by frequent occlusions, cluttered background and crowded scenes which significantly degrade performance. Yet, strategies leveraging the complementarity of multiple shots and views from different video streams is overlooked.
In this paper, we study strategies for an end-to-end pipeline that operates from multiple video streams to detect and re-identify luggage.
We evaluate different variants on a benchmark dataset acquired in real-world environments to address the questions of selecting and merging different views and shots of the same luggage as well as training on different environment datasets for an overall ReID performance improvement in practical applications.
%    Luggage re-identification (ReID) is a critical task in video surveillance, enabling the tracking of objects across disjoint camera views in complex environments such as airports and train stations. While existing approaches primarily focus on appearance-based matching under relatively clean conditions, real-world scenarios are characterized by frequent occlusions, crowded scenes, and fragmented trajectories, which significantly degrade performance.
    %In this paper, we propose an end-to-end pipeline for luggage re-identification that operates directly from raw video streams and is robust to severe occlusions. The proposed framework integrates detection, multi-object tracking, occlusion-aware feature extraction, and cross-camera association into a unified system. We evaluate our approach on a custom benchmark dataset. Experimental results demonstrate that our pipeline consistently performs remarkable scores in challenging occluded scenarios, achieving significant improvements in re-identification accuracy and robustness. These results highlight the importance of holistic, pipeline-level design for practical deployment of luggage ReID systems in real-world surveillance applications.
\end{abstract}


\section{Introduction}

Luggage re-identification (ReID) has emerged as a critical capability in modern video surveillance systems, particularly in large-scale environments such as airports and transportation hubs. Existing approaches typically focus on improving feature extraction or metric learning strategies for ReID, often assuming clean detections and fully visible objects. Yet, practical deployment raises several system-level questions that remain largely unexplored: How many cameras are actually required to achieve reliable luggage ReID? How many observations should be gathered for each tracked instance? Does aggregating all available frames improve robustness, or can noisy detections degrade performance? How should multi-camera information be exploited in asymmetric surveillance setups?
%The ability to associate a piece of luggage across disjoint camera networks enables applications ranging from lost item retrieval to security monitoring. 
%However, unlike person ReID, luggage ReID remains relatively underexplored and presents unique challenges due to its limited discriminative features and high susceptibility to occlusions.

To address these questions, we propose an end-to-end pipeline for luggage ReID across multiple scenes in a realistic asymmetric surveillance setting. The gallery' scene corresponds to the enrollment stage, where luggage items are observed in a controlled environment, e.g., moving alone on a conveyor belt. In contrast, the query' scene represents unconstrained surveillance areas, where luggage is carried by people and may be partially occluded, viewed from limited angles, or only intermittently visible. Both scenes include multiple cameras providing complementary viewpoints, enabling the proposed approach to leverage multi-view observations rather than relying on single-frame appearance features. This multi-camera setup helps mitigate occlusions and improves the robustness of cross-scene matching.
Such a framework is particularly relevant for applications including the retrieval of lost or abandoned luggage, the ReID of suspicious bags previously scanned by security systems, and forensic analysis in large transportation hubs. This practical context further motivates the need for robust multi-camera and multi-view ReID approaches capable of operating under realistic surveillance conditions.
%
%The proposed pipeline integrates object detection, multi-object tracking, and feature aggregation into a unified framework. By leveraging multiple observations of the same luggage across cameras, the system constructs robust representations that remain discriminative even under severe partial occlusions. Cross-scene matching is then performed between aggregated representations from the query and gallery scenes.

The main contributions of this work are as follows:
(1) we propose a practical end-to-end multi-camera pipeline for luggage ReID in realistic surveillance environments, explicitly designed for asymmetric enrollment-vs-monitoring scenarios;
(2) we introduce a multi-view and temporal aggregation strategy adapted to the asymmetric nature of our setup, improving robustness to occlusions, missed detections, and fragmented observations;
(3) we conduct extensive experimental and ablation studies, including analyses of camera configurations, shot selection strategies, aggregation mechanism, and cross-dataset adaptation, providing practical insights for real-world deployment of luggage ReID systems.

\section{Related Work}

\subsection{Object and Luggage Re-Identification}

ReID aims at matching instances of the same object across different camera views in non-overlapping surveillance environments. While person, animal and vehicle ReID ~\cite{zheng2016person, ye2021deep, kuma2019vehicle} have been extensively studied, luggage ReID has received comparatively less attention despite its importance in transportation and security applications~\cite{yang2022quadnet, liu2025cdr, zhao2025luggage, mazzeo2020siamese, wang2020model}.
Existing luggage ReID methods~\cite{yang2022quadnet, liu2025cdr, zhao2025luggage, mazzeo2020siamese, wang2020model} primarily focus on learning discriminative appearance features from cropped image data. For example, QuadNet \cite{yang2022quadnet} introduces a quadruplet loss to learn multi-view representations that are robust to viewpoint variations. However, these methods typically operate on pre-cropped images and assume relatively clean detections, limiting their applicability in real surveillance scenarios.

\subsection{Whole ReID Pipelines for Real-World Applications}

Recent advances in multi-object tracking~\cite{leal2015motchallenge} have incorporated ReID features to improve identity association over time in the same camera. Tracking-by-detection frameworks such as DeepSORT \cite{wojke2017simple} and variants \cite{bewley2016simple,zhang2022bytetrack} combine motion cues with appearance embeddings to maintain identity consistency across frames.
More recent unified approaches jointly optimize detection and ReID within a single framework. While these methods \cite{wang2020towards,zhang2021fairmot} improve tracking robustness, they are primarily designed for single-camera settings and do not explicitly address cross-scene or multi-camera luggage ReID. 
Also,
%Moreover, 
they often assume relatively continuous object visibility, which is not guaranteed in real-world surveillance environments.
Beyond academic research, several industrial solutions have recently explored AI-based luggage ReID systems for real-world airport operations. For instance, a commercial solution automates the ReID of lost or untagged luggage using multi-view image acquisition and visual matching techniques in airport baggage handling systems \cite{alix}. Yet, unlike our asymmetric scenario, both enrollment and query acquisitions are under controlled environments and luggage is isolated. Such systems further highlight the operational importance of robust luggage ReID pipelines for applications including lost luggage retrieval, baggage tracking, and security investigations.

{\captionsetup{justification=centering}
\begin{figure*}[t]
    \centering
    \includegraphics[width=0.75\textwidth]{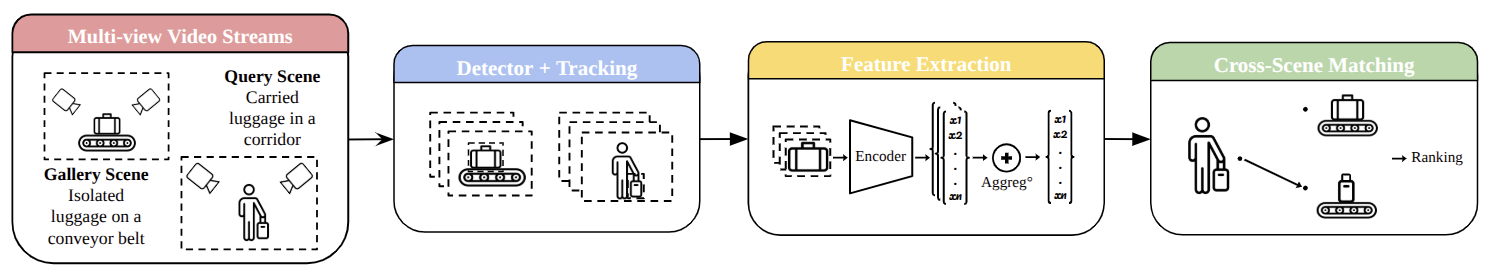}
    \caption{ReID Pipeline Overview}
    \label{fig:pipeline}
\end{figure*}
}

\subsection{Occlusion and Multi-View Robust Re-Identification}

Occlusion remains one of the most challenging problems in ReID systems. Existing works have proposed part-based representations \cite{sun2018beyond}, attention mechanisms \cite{xu2018attention}, and temporal modeling strategies \cite{liu2017video} to mitigate the effect of missing or partially visible information. Other approaches specifically target occluded ReID through pose-guided alignment or part discovery strategies \cite{miao2019pose,li2021diverse}. However, most of these approaches focus on person ReID and assume relatively structured motion patterns.

Multi-view learning approaches attempt to leverage complementary camera viewpoints to improve robustness \cite{lin2019multi,zhang2019mvb,yang2022quadnet}. These methods demonstrate that combining information across views can significantly enhance feature discriminability. Yet, they typically rely on synchronized or fully observed multi-view data, and do not consider real asymmetric surveillance scenarios where non-cooperative behaviors may lead to fragmented or incomplete observations.

\section{Method}

We consider the problem of luggage ReID across two disjoint scenes with distinct characteristics: a "query" scene and a "gallery" scene. Each scene is monitored by multiple cameras providing different viewpoints.

In the gallery scene, luggage is transported on a conveyor belt, resulting in predictable and consistent motion patterns. This controlled setting allows trajectories to be reliably associated across cameras, enabling the extraction of representative observations of each luggage instance. In contrast, the query scene corresponds to a more complex environment where luggage is carried by individuals, leading to irregular motion, partial visibility, and fragmented trajectories that cannot be consistently associated across cameras.

Given a luggage instance enrolled in the gallery scene, the objective is to retrieve the corresponding observations in the query scene across realistic surveillance conditions, enabling practical applications such as lost luggage retrieval, suspicious baggage search, or forensic analysis.
%Given a set of observations of a target luggage in the query scene, the objective is to identify the corresponding instance in the gallery scene.

\subsection{Pipeline Overview}

We propose a pipeline for luggage ReID across disjoint scenes, explicitly designed to handle the asymmetric characteristics of setup.
%The pipeline processes a gallery scene and a query scene, each observed by multiple cameras, and produces robust representations for cross-scene matching.
An overview of the proposed framework is illustrated in Fig.~\ref{fig:pipeline}. The pipeline consists of four main stages:

\subsubsection{Detection and Tracking}

Each video stream is processed independently to detect luggage instances and then associated to form tracks over time. In the gallery scene, the structured motion of luggage allows consistent association of tracks across cameras. In contrast, in the query scene, tracking is performed per camera due to the absence of reliable cross-camera associations.

\subsubsection{Scene-Specific Representation Construction}

%The pipeline constructs representations differently for the gallery and query scenes because of the asymmetric nature of the problem.
%In the gallery scene, one representative observation is selected per camera, resulting in a compact multi-view description of each luggage instance.
%In the query scene, multiple observations within each track are aggregated over time to produce a robust feature representation.

Due to the asymmetric nature of the considered surveillance setup, several strategies can be considered to construct feature representations for the gallery and query scenes. Depending on the acquisition conditions, a luggage instance may be represented using a single observation, multiple observations from different cameras, or multiple temporal observations extracted from a track.
In controlled environments where trajectories are predictable and objects are consistently observed across viewpoints, multi-camera representations can be constructed by selecting one or several observations per camera. In less constrained surveillance areas, where visibility may be fragmented due to occlusions or unconstrained motion, multiple observations can instead be extracted from a track and combined into a track-level representation.

\subsubsection{Feature Extraction}

For each detected instance, visual features are extracted using a deep neural network. These features capture appearance cues that should be invariant to viewpoint and illumination changes, and can then be aggregated depending on the multi-shot aggregation strategy.

\subsubsection{Cross-Scene Matching}

Finally, ReID is performed by matching query representations against gallery representations. Each query track is compared to all gallery tracks using a similarity metric, and gallery candidates are ranked accordingly.

This design leverages the complementary strengths of both scenes: structured multi-view information in the gallery and temporal redundancy in the query. As a result, the proposed pipeline effectively mitigates the impact of occlusions and viewpoint variations in realistic surveillance scenarios.

\subsection{Detection and Multi-Object Tracking}

The first stage of the proposed pipeline, 
applied independently to each camera stream, 
consists of detecting luggage instances and associating them over time to form tracks. 
%This stage is applied independently to each camera stream.

\subsubsection{Detection}

Luggage instances are detected in each frame using an object detector that provides bounding boxes (BBox) and confidence scores $s_c$ for each candidate object. Only detections with $s_c$ above a predefined threshold $\theta_c$ are retained to reduce false positives.
Then, the asymmetric tracking strategy reflects realistic surveillance conditions, where controlled environments (e.g., conveyor systems) allow multi-view association, while unconstrained environments lead to fragmented and incomplete tracks. This difference motivates the use of scene-specific representation strategies in the subsequent stages of the pipeline.

\subsubsection{Tracking in the Gallery Scene}

The gallery scene exhibits a well-controlled environment in which luggage moves on a conveyor belt, at a controlled and constant speed. This results in predictable and consistent motion across cameras, free from any obstruction, enabling reliable association of trajectories between different views.

Leveraging this property, detections corresponding to the same luggage instance are associated across cameras based on spatio-temporal consistency. This allows the construction of cross-camera tracks, where each luggage instance is observed sequentially by multiple cameras.

%For each trajectory, a single representative observation is selected per camera, resulting in a fixed set of multi-view samples for each luggage instance.

\subsubsection{Tracking in the Query Scene}

In contrast, in the query scene, luggage is carried by people, resulting in less predictable motion patterns, frequent occlusions, and partial visibility across cameras. As a consequence, reliable cross-camera association is not feasible.

Tracking is therefore performed independently for each camera using a tracking-by-detection approach, where detections are associated across consecutive frames based on spatial proximity and appearance similarity. This results in a set of per-camera tracks, each corresponding to a sequence of observations of a luggage instance over time. Due to occlusions and missed detections, these tracks may be fragmented.

\subsection{Feature Extraction and Aggregation}

Due to the unconstrained nature of the query scene, individual observations of a luggage instance are often affected by occlusions, motion blur, and imperfect detections. As a result, directly aggregating all observations may introduce noise and degrade the quality of the resulting multi-view and multi-shot representation.
For each detected instance, we first extract a visual feature representation using a deep convolutional neural network. More precisely, given a BBox, the corresponding image region is cropped, resized to a fixed resolution, and passed through the network to obtain a feature vector $f \in \mathbb{R}^d$, where $d$ is the output dimension of the network. 
%The feature extractor is initialized from a model pretrained on a large-scale dataset and fine-tuned on a custom constructed luggage dataset.

To construct robust query representations, we aggregate features at the track level. Given a track $\tau = {x_1, x_2, \dots, x_T}$, where $x_i$ is the $i^{th}$ detected instance of a track and $T$ is the number of instances in this track, we first discard unreliable detections based on their confidence scores so that the remaining satisfy $s_{c_i} \geq \theta_c$.% Specifically, only observations with detection confidence above a threshold of 0.7 are retained. This filtering step removes low-quality or ambiguous detections that are more likely to correspond to occluded or poorly localized instances.

From the remaining observations, we extract feature vectors $({f_1, f_2, \dots, f_K})$, where $K \leq T$. The final multi-shot query representation $F_q$ is then computed as the average of these features:
$
%\[
F_q = \frac{1}{K} \sum_{i=1}^{K} f_i
%\]
$

This simple aggregation strategy leverages temporal redundancy within trajectories while ensuring that only reliable observations contribute to the final representation. By filtering out low-confidence detections prior to aggregation, the method reduces the impact of occlusions and noise without requiring additional weighting mechanisms.

In contrast, the gallery scene provides structured multi-view observations with limited occlusion. Therefore, no temporal aggregation is required, and each gallery instance $\gamma$ is composed of one shot from each camera, such as $\gamma = x_1, x_2, \dots, x_{\#GCams}$, where $\#GCams$ is the number of cameras in the gallery scene and $x_i$ is the detected instance of the $i^{th}$ gallery camera. %and each gallery observation is treated independently during matching.

\subsection{Cross-Scene Matching}

The final stage of the pipeline consists of matching query representations against gallery representations to identify corresponding luggage instances across scenes.
Given a query track, we obtain a single aggregated feature vector $F_q$ using the proposed temporal aggregation strategy. In contrast, each gallery instance is represented by a set of multi-view features ({$F_g^1$, $F_g^2$, \dots, $F_g^{\#GCams}$}).
During matching, we compute the distance between the query feature $F_q$ and each gallery feature independently. This results in a set of distances between the query and all $\#GCams$ gallery features.
All gallery features are then ranked in ascending order of their distance to the query. Since multiple gallery features may correspond to the same luggage ID (by multi-view association), the final ranking at the identity level is obtained by considering the best-ranked occurrence of each ID.
This formulation allows the matching process to naturally exploit complementary viewpoints in the gallery: even if some observations are less informative or affected by occlusions, other views of the same luggage instance can still yield a low distance and lead to correct identification.
%
%This approach aligns with realistic deployment conditions, where gallery observations are processed independently and no explicit fusion is required at matching time.

\section{Experiments}

\subsection{Dataset Construction}

To evaluate the proposed approach under realistic surveillance conditions, we construct a multi-camera luggage ReID dataset from video footage from a real airport, in the customs area (cf. Fig.~\ref{fig:results}). The dataset consists of two disjoint scenes, referred to as the gallery and query scenes, each captured by multiple cameras with different viewpoints.

While the gallery scene captures luggage items moving on a conveyor belt immediately after plane landing, the query scene corresponds to passengers leaving the customs area through a corridor while carrying their luggage

The acquisition system consists of a multi-camera configuration with 5 cameras in the gallery scene and 6 cameras in the query scene. In both environments, cameras are positioned around the area of interest and oriented toward a common region to maximize coverage of luggage instances.
%This setup ensures that objects are observed from multiple viewpoints in the gallery scene, while providing diverse and potentially fragmented observations in the query scene due to unconstrained motion.
To establish cross-scene correspondences, we manually annotate matching luggage instances between the query and gallery scenes by selecting consistent trajectories.

%Luggage instances are first detected using an object detector with publicly available pretrained weights. The detections are then associated over time to form trajectories within each camera.

The resulting dataset contains a total of 116 bags, with multiple observations of each luggage instance across different cameras and time steps, naturally including variations in viewpoint, illumination, and partial occlusions. Overall, the dataset contains 579 gallery tracks and 745 query tracks, with approximately 18 images available per track on average. This setup enables the evaluation of multi-view aggregation strategies in realistic conditions.

%\begin{table}[!h]%[!b]
%    \centering
%    \begin{tabular}{l | ccc}
%         & Bags & Tracks & Images per track\\
%        \hline
%        Gallery & 116 & 579 & 1 \\
%        Query & 116 & 745 & \~ 18 \\
%    \end{tabular}
%    \\ [1ex]
%    \caption{Dataset details}
%    \label{tab:dataset}
%\end{table}

\subsection{Implementation Details}

Object detection is performed using the CoDETR detector \cite{zong2023detrs} with publicly available pretrained weights. This object detector was selected for its strong general object detection performance and its ability to handle complex scenes with multiple overlapping objects.%The detector is applied independently to each video stream to extract bounding boxes corresponding to luggage instances.
%
%cite bytetrack
%Multi-object tracking is then conducted on each camera using a tracking-by-detection approach, where detections are associated across frames based on spatial proximity and appearance similarity to form trajectories. These trajectories serve as the basic units for subsequent feature aggregation.
%
%In the gallery scene, luggage instances move along a conveyor belt, resulting in predictable trajectories. This allows reliable association of detections across cameras, enabling the construction of cross-camera trajectories.
For each luggage instance of the gallery scene, one representative frame is selected per camera by choosing the observation where the luggage instance is closest to the center of the field of view, resulting in a fixed set of 5 images (corresponding to the 5 cameras) that form a structured multi-view representation.
%
%In contrast, the query scene exhibits unconstrained human motion, making cross-camera trajectory association unreliable. Therefore, tracking is performed independently per camera, and trajectories are not merged across views.
For each query track, feature embeddings are extracted and aggregated over time using the proposed aggregation strategy.
For each detected instance, visual features are extracted using the TransReID algorithm \cite{he2021transreid}  that is pretrained on the large-scale luggage dataset MVB \cite{zhang2019mvb} and fine-tuned on our constructed dataset. Input images are resized to a fixed resolution of 256*256 pixels, and standard data augmentation techniques are applied during training \cite{luo2019bag, mclaughlin2015data}.
During inference, each query track is represented by a single aggregated feature vector, while each gallery instance is represented by a set of multi-view features. 
%Similarity between query and gallery representations is computed using Euclidean distance.
Query and gallery representations are compared with Euclidean distance.

\subsection{Results}

We evaluate the proposed approach using standard ReID metrics, including cumulative matching characteristics (CMC) at rank-1, rank-5, and rank-10, as well as mean Average Precision (mAP). To provide a comprehensive assessment, we adopt a bidirectional evaluation protocol and report results for both query-to-gallery (Q→G), for the recognition of every query bag, and gallery-to-query (G→Q) matching, for the recall of every gallery bag.

To assess the contribution of each component of the pipeline, we compare the proposed method with several configurations. First, we consider a random single-frame baseline, where one observation per track is selected uniformly at random. To ensure stability, results for this baseline are averaged over ten runs. Second, we evaluate a score-based single-frame variant, where the observation with the highest detection confidence is selected for each track. Third, we consider a variant that uses all available observations per track without applying any confidence-based filtering, followed by feature aggregation. Lastly, we show the results of the proposed approach, that is when considering only the $K$ most confident detections, based on detection scores, and aggregating them. Here $K=5$, an extensive study on this parameter is presented shortly after.

% Results & baseline
% Our best result is K=5-shots with det score criterion

\begin{table}[!h]%[!b]
    \centering
    \scalebox{0.9}{
        \begin{tabular}{l ll | cccc}
             \hline
              & & & \multicolumn{4}{c}{\textbf{Q$\rightarrow$G}} \\
             \hline
             Variant & K & Conf. & R@1 & R@5 & R@10 & mAP\\
             \hline
             Random frame & 1 & - & 0.742 & 0.861 & 0.903 & 0.732$\pm$ 0.007 \\
             Most confident & 1 & \checkmark & 0.794 & 0.899 & \textbf{0.935} & 0.775 \\
             All frames & all & - & 0.792 & 0.874 & 0.914 & 0.769 \\
             K-most confident & 5 & \checkmark & \textbf{0.812} & \textbf{0.903} & 0.931 & \textbf{0.784} \\
             \hline
              & & & \multicolumn{4}{c}{\textbf{G$\rightarrow$Q}} \\
             \hline
             Variant & K & Conf. & R@1 & R@5 & R@10 & mAP\\
             \hline
             Random frame & 1 & - & 0.784 & 0.904 & 0.931 & 0.705$\pm$ 0.005 \\
             Most confident & 1 & \checkmark & 0.826 & \textbf{0.924} & \textbf{0.963} & 0.757 \\
             All frames & all & - & 0.831 & 0.917 & 0.934 & 0.756 \\
             K-most confident & 5 & \checkmark & \textbf{0.837} & 0.921 & 0.942 & \textbf{0.767} \\
             \hline
        \end{tabular}
    }
    \\ [1ex]
    \caption{Variants comparison results}
    \label{tab:results}
\end{table}

As shown in Tab.~\ref{tab:results}, the random single-frame baseline yields the lowest performance, highlighting the limitations of relying on arbitrary observations. Selecting the highest-confidence frame improves results, confirming that detection confidence is a reliable indicator of observation quality even for ReID.

{\captionsetup{justification=centering}
\begin{figure}[h]
    \centering
    \includegraphics[width=0.6\columnwidth]{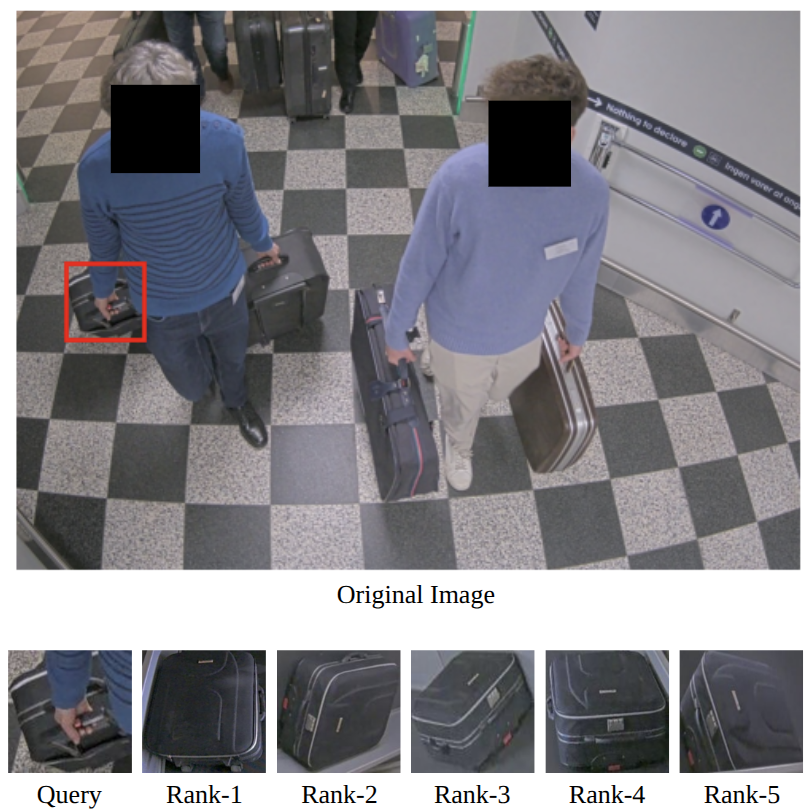}
    \caption{Visualization of a query with ReID ranking}
    \label{fig:results}
\end{figure}
}

Using all frames without filtering further degrades performance compared to the score-based baseline, indicating that low-confidence detections introduce noise into the aggregated representation. In contrast, the proposed approach, which combines confidence-based filtering with feature aggregation, achieves the best performance. This demonstrates that removing unreliable observations prior to aggregation is critical for constructing robust representations.

We observe that the score-based single-frame baseline achieves slightly higher rank-10 performance than the proposed method. This may be because selecting the highest-confidence frame preserves a particularly discriminative observation that still appears among the top-ranked candidates, even if not ranked first. However, the proposed approach consistently outperforms the baseline in rank-1 and mAP, which are more important metrics for practical deployment.

In real-world applications, ReID systems are often integrated into automated pipelines where decisions rely on the top-ranked match rather than a shortlist of candidates. In this context, improving rank-1 accuracy is essential, as it directly impacts the reliability of automatic identification. Similarly, higher mAP reflects better overall ranking quality across all queries. Therefore, despite slightly lower rank-10 performance, the proposed method provides a more robust and practically relevant solution.

Overall, these results validate the design choices of the proposed pipeline and emphasize the importance of both filtering and aggregation for handling noisy and partially occluded observations in realistic surveillance scenarios.

Lastly, all experiments were conducted on a single NVIDIA TITAN RTX GPU. 
Inference time of detection and tracking is 6.1s for the synchronized images of one scene's cameras whereas ReID runtime is 30ms for a single query.

%In Tab.~\ref{tab:inference_cost}, we analyze the inference cost of detection and tracking separately from feature extraction, aggregation, and ReID, since these stages have significantly different computational costs. The detection and tracking runtime is measured on the synchronized images of one scene's cameras, whereas ReID runtime is measured on a single query.

%\begin{table}[h]%[!b]
%    \centering
%    \scalebox{0.8}{
%        \begin{tabular}{l | cc}
%             \hline
%              & Detection \& Tracking & ReID \\
%             \hline
%             Inference cost (s) & 6.10 & 0.03 \\
%             \hline
%        \end{tabular}
%    }
%    \\ [0.5ex]
%    \caption{Inference cost analysis}
%    \label{tab:inference_cost}
%\end{table}

\subsection{Ablation Study}

%List of experiments to show:\\
%default set-up = no aggregation, best detection for each query track\\
%feature aggregation : across same track @query\\
%camera study : all cams, one cam @query, one cam @gallery, duo best cams\\
%k-shot study : WITH feature aggregation do 1,2,3,max shots per track @query\\
%k-means study : with k cluster centers

We conduct a series of ablation experiments to evaluate the contribution of each component of the proposed pipeline, including multi-camera configurations, temporal aggregation strategies, and dataset characteristics.

\subsubsection{Impact of Multi-Camera Configurations}

We first investigate the impact of the number and selection of cameras in both query and gallery scenes. To this end, we evaluate different configurations, including single-camera setups, camera pairs, and the full multi-camera setting.

To simplify the presentation of the different camera configurations, we introduce the following notation. Individual cameras are identified based on their relative position in the scene. For instance, $GCam_{FR}$ refers to the gallery camera located at the front-right of the scene, while similar notation is used for other viewpoints (e.g., front-left, lateral-left, lateral-right, rear-left, rear-right). The same convention applies to the query scene cameras (front-left, front-right, lateral-left, lateral-right, rear). We denote $B_Q$ and $B_G$ the number of bags detected by the query/gallery scenes' cameras in a specific configuration, while $\#QCams$ and $\#GCams$ refer to the number of cameras used for an experiment.

% Camera study results
\begin{table}[h]%[!b]
    \centering
    \scalebox{0.8}{
        \begin{tabular}{lll | cccc}
             \hline
              $\#GCams=5$ & & & \multicolumn{4}{c}{\textbf{Q$\rightarrow$G}} \\
             \hline
             Camera & B$_G$ & B$_Q$ & R@1 & R@5 & R@10 & mAP\\
             \hline
             QCam$_{FL}$ & 116 & 100  & 0.820 & 0.900 & 0.930 & 0.792 \\
             QCam$_{FR}$ & 116 & 104  & 0.798 & 0.933 & 0.962 & 0.792 \\
             QCam$_{LL}$ & 116 & 106  & 0.783 & 0.868 & 0.906 & 0.771 \\
             QCam$_{LR}$ & 116 & 110  & 0.791 & 0.882 & 0.945 & 0.775 \\
             QCam$_{RL}$ & 116 & 98  & 0.776 & 0.878 & 0.908 & 0.739 \\
             QCam$_{RR}$ & 116 & 97  & 0.794 & 0.928 & 0.938 & 0.773 \\
             \hline
             $\#GCams=5$ & & & \multicolumn{4}{c}{\textbf{G$\rightarrow$Q}} \\
             \hline
             Camera & B$_G$ & B$_Q$ & R@1 & R@5 & R@10 & mAP\\
             \hline
             QCam$_{FL}$ & 116 & 100  & 0.661 & 0.789 & 0.812 & 0.720 \\
             QCam$_{FR}$ & 116 & 104  & 0.684 & 0.839 & 0.870 & 0.756 \\
             QCam$_{LL}$ & 116 & 106  & 0.680 & 0.822 & 0.853 & 0.749 \\
             QCam$_{LR}$ & 116 & 110  & 0.710 & 0.879 & 0.912 & 0.787 \\
             QCam$_{RL}$ & 116 & 98  & 0.617 & 0.737 & 0.788 & 0.674 \\
             QCam$_{RR}$ & 116 & 97  & 0.611 & 0.765 & 0.794 & 0.683 \\
             \hline
        \end{tabular}
    }
    \\ [1ex]
    \caption{Single camera results @Query : $\#QCams=1$ 
    %$\#GCams=5$
    }
    \label{tab:single_cam_query}
\end{table}

\begin{table}[h]%[!b]
    \centering
    \scalebox{0.8}{
        \begin{tabular}{lll | cccc}
             \hline
              $\#QCams=6$ & & & \multicolumn{4}{c}{\textbf{Q$\rightarrow$G}} \\
             \hline
             Camera & B$_G$ & B$_Q$ & R@1 & R@5 & R@10 & mAP\\
             \hline
             GCam$_{FL}$ & 116 & 114  & 0.761 & 0.917 & 0.948 & 0.834 \\
             GCam$_{FR}$ & 116 & 114  & 0.763 & 0.915 & 0.953 & 0.834 \\
             GCam$_{LL}$ & 116 & 114  & 0.714 & 0.896 & 0.933 & 0.797 \\
             GCam$_{LR}$ & 116 & 114  & 0.746 & 0.899 & 0.943 & 0.819 \\
             GCam$_{R}$ & 115 & 114  & 0.714 & 0.899 & 0.953 & 0.799 \\
             \hline
              $\#QCams=6$  & & & \multicolumn{4}{c}{\textbf{G$\rightarrow$Q}} \\
             \hline
             Camera & B$_G$ & B$_Q$ & R@1 & R@5 & R@10 & mAP\\
             \hline
             GCam$_{FL}$ & 116 & 114  & 0.828 & 0.940 & 0.966 & 0.770 \\
             GCam$_{FR}$ & 116 & 114  & 0.853 & 0.940 & 0.974 & 0.771 \\
             GCam$_{LL}$ & 116 & 114  & 0.793 & 0.888 & 0.957 & 0.737 \\
             GCam$_{LR}$ & 116 & 114  & 0.828 & 0.922 & 0.957 & 0.753 \\
             GCam$_{R}$ & 115 & 114  & 0.835 & 0.939 & 0.965 & 0.751 \\
             \hline
        \end{tabular}
    }
    \\ [1ex]
    \caption{Single camera results @Gallery : %$\#QCams=6$ 
    $\#GCams=1$}
    \label{tab:single_cam_gallery}
\end{table}

Results show that performance varies significantly across cameras, indicating that some viewpoints provide more discriminative information than others. Single-camera configurations generally reduce performance due to limited visibility and missed detections. In particular, Tab.~\ref{tab:single_cam_query} highlights the asymmetric impact of camera setups in query and gallery scenes. Using a single query camera can still provide relatively strong query-to-gallery performance, as some viewpoints remain discriminative, but it noticeably degrades gallery-to-query results because of missed detections and incomplete luggage coverage. Fig.~\ref{fig:cameras_comparison} illustrates ReID errors with a single camera, while combining all cameras yields correct matches. Similarly, restricting the gallery to one camera significantly lowers query-to-gallery performance, especially rank-1 accuracy (Tab.~\ref{tab:single_cam_gallery}), since limited gallery views reduce the chances of capturing discriminative observations.

{\captionsetup{justification=centering}
    \begin{figure}[h]
        \centering
        \includegraphics[width=0.7\columnwidth]{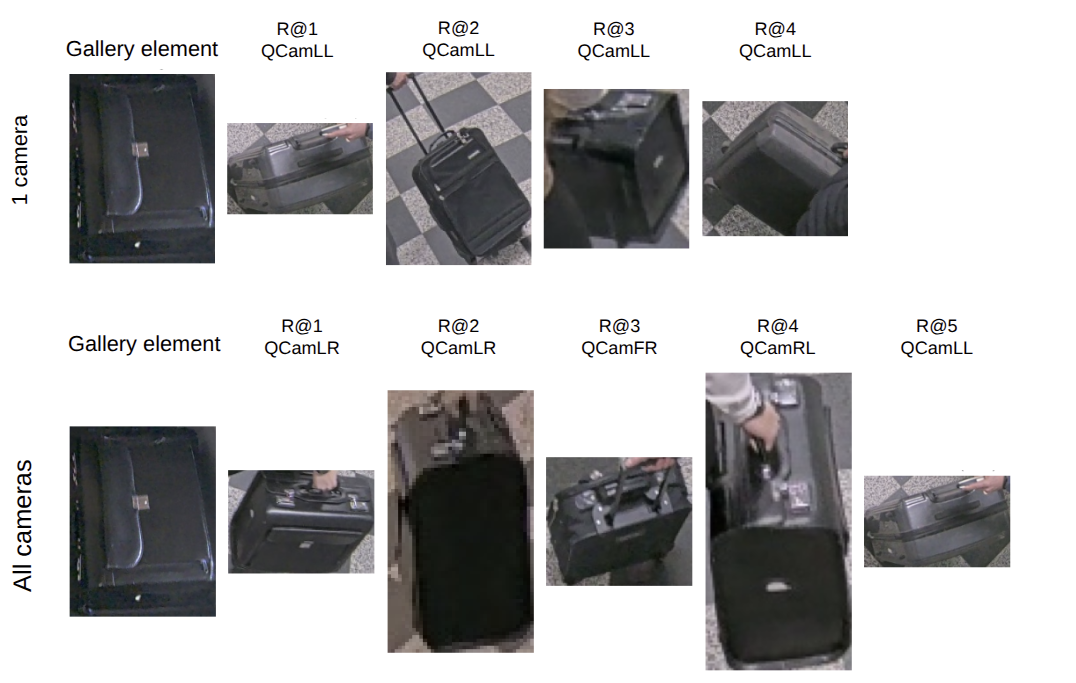}
        \caption{
        Comparison of recall by one camera vs. all cameras
        %Comparison of recall with one camera and with all cameras
        }
        \label{fig:cameras_comparison}
    \end{figure}
}

On a side note, Tab.~\ref{tab:single_cam_gallery} also reveals that when considering all available query cameras, some luggage instances remain undetected. This is primarily due to low detection confidence in challenging situations, such as bags piled together on trolleys, which leads to missed or unreliable detections. This observation accentuates the inherent difficulty of the task and the impact of detection quality on overall ReID performance.

These findings highlight the importance of multi-camera setups in realistic surveillance environments, where individual cameras may fail to capture complete object information due to occlusions or limited fields of view.

\subsubsection{Impact of Temporal Aggregation}

We next analyze the effect of temporal aggregation in the query scene. Specifically, we vary the number of observations ($K$-shot) used to construct track-level representations, and compare the impact of having an aggregation or not. We consider different selection strategies, including choosing the top-$K$ observations based on detection confidence or BBox size.

% K-shots results
\begin{table}[h]%[!b]
    \centering
    \scalebox{0.7}{
        \begin{tabular}{l  l  l | cccc | cccc}
             \hline
              & & & \multicolumn{4}{c|}{\textbf{Q$\rightarrow$G}} & \multicolumn{4}{c}{\textbf{G$\rightarrow$Q}} \\
             \hline
             K & Aggreg. & Selection & R@1 & R@5 & R@10 & mAP & R@1 & R@5 & R@10 & mAP\\
             \hline
             1 & $\varnothing$ & score & 0.794 & 0.899 & 0.935 & 0.775 & 0.826 & 0.924 & \textbf{0.963} & 0.757 \\
             1 & $\varnothing$ & size & 0.782 & 0.883 & 0.913 & 0.768 & 0.812 & 0.917 & 0.947 & 0.748 \\
             \hline
             2 & no & score & 0.782 & 0.890 & 0.924 & 0.763 & 0.831 & 0.902 & 0.935 & 0.733 \\
             2 & no & size & 0.783 & 0.893 & 0.922 & 0.771 & 0.800 & 0.910 & 0.944 & 0.739 \\
             2 & yes & score & 0.816 & 0.901 & 0.936 & 0.785 & 0.814 & 0.924 & 0.949 & 0.755 \\
             2 & yes & size & 0.799 & 0.896 & 0.934 & 0.782 & 0.812 & 0.924 & 0.953 & 0.763 \\
             \hline
             3 & no & score & 0.779	& 0.887 & 0.925 & 0.761 & 0.824 & 0.893 & 0.921 & 0.724 \\
             3 & no & size & 0.791 & 0.891 & 0.924 & 0.778 & 0.821 & 0.902 & 0.930 & 0.738 \\
             3 & yes & score & 0.804 & 0.903 & 0.937 & 0.784 & 0.817 & 0.921 & 0.946 & 0.761 \\
             3 & yes & size & 0.809 & 0.906 & 0.934 & \textbf{0.796} & 0.822 & \textbf{0.937} & 0.954 & \textbf{0.773} \\
             \hline
             4 & no & score & 0.778 & 0.890 & 0.925 & 0.760 & 0.822 & 0.905 & 0.923 & 0.719 \\
             4 & no & size & 0.789 & 0.891 & 0.923 & 0.776 & 0.819 & 0.907 & 0.926 & 0.734 \\
             4 & yes & score & 0.807 & 0.898 & 0.930 & 0.781 & 0.815 & 0.926 & 0.947 & 0.759 \\
             4 & yes & size & 0.812 & 0.906 & 0.939 & \textbf{0.796} & 0.814 & 0.931 & 0.951 & 0.772 \\
             \hline
             5 & no & score & 0.777 & 0.892 & 0.926 & 0.761 & 0.828 & 0.896 & 0.919 & 0.718 \\
             5 & no & size & 0.788 & 0.889 & 0.921 & 0.773 & 0.819 & 0.910 & 0.924 & 0.730 \\
             5 & yes & score & 0.812 & 0.903 & 0.931 & 0.784 & \textbf{0.837} & 0.921 & 0.942 & 0.767 \\
             5 & yes & size & 0.815 & \textbf{0.908} & \textbf{0.941} & 0.794 & 0.798 & 0.924 & 0.949 & 0.771 \\
             \hline
             all & no & all & 0.772 & 0.882 & 0.920 & 0.755 & 0.835 & 0.903 & 0.919 & 0.686 \\
             all & yes & all & \textbf{0.819} & 0.897 & 0.929 & 0.789 & 0.819 & 0.910 & 0.946 & 0.761 \\
             \hline
        \end{tabular}
    }
    \\ [1ex]
    \caption{K-shot study}
    \label{tab:k_shot}
\end{table}

%The results demonstrate that aggregation significantly improves performance compared to single-frame representations, confirming the benefit of leveraging temporal redundancy. Increasing the number of observations generally leads to better performance up to a certain point, after which gains saturate.

When observations are selected according to the detection confidence score, Tab.~\ref{tab:k_shot} shows that increasing $K$ without feature aggregation gradually decreases performance, suggesting that lower-quality observations introduce noise. In contrast, feature aggregation consistently improves performance for the same $K$, as averaging multiple observations yields more robust descriptors. However, using all available observations significantly increases inference time, highlighting a trade-off between accuracy and computational cost.

The same trends are observed when using BBox size as the selection criterion, although performance at low $K$ is generally lower than with detection confidence. This difference becomes less pronounced for larger $K$, where size-based selection can achieve comparable or slightly better results. This suggests that detection confidence is better for selecting a few reliable observations, while size-based selection may retain more complementary appearance information for larger subsets.
%
%We also observe that 
Also,
selecting observations based on detection confidence provides a good trade-off between performance and robustness. This validates the design choice of filtering low-confidence detections and averaging the remaining features in our method.

\subsubsection{Dataset and Domain Analysis}

We analyze the impact of dataset characteristics and domain adaptation using 3 datasets: a large public dataset (MVB), our primary custom dataset (CD-1), and a second dataset acquired in a different location with a similar setup (CD-2). Both datasets are split for train (\textit{-tr}) and test (\textit{-te}) purposes. We observe that CD-2 appears to be an easier dataset for the ReID task, likely due to its more uniform background (e.g., consistent floor coloring), which reduces visual clutter and may also correlate with more stable lighting conditions and fewer distractors compared to CD-1.

{\captionsetup{justification=centering}
\begin{figure}[h]
    \centering
    \includegraphics[width=0.7\columnwidth]{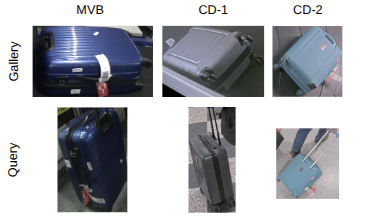}
    \caption{Sample of images from each dataset}
    \label{fig:datasets}
\end{figure}
}

We first evaluate on CD-1-te a model pretrained on MVB. The results from Tab.~\ref{tab:domain_analysis} show limited performance, indicating a significant domain gap between the external dataset and the target environment. Fine-tuning the same model on CD-1-tr leads to substantial improvement, confirming the importance of adapting feature representation to target domain.

\begin{table}[h]%[!b]
    \centering
    \scalebox{0.8}{
        \begin{tabular}{ll | cccc}
             \hline
             Model used & & \multicolumn{4}{c}{\textbf{Q$\rightarrow$G}} \\
             \hline
             Training & Test & R@1 & R@5 & R@10 & mAP\\
             \hline
             MVB & CD-1-te & 0.611 & 0.797 & 0.846 & 0.559 \\
             MVB + finetuned on CD-1-tr & CD-1-te & 0.794 & 0.899 & 0.935 & 0.775 \\
             MVB & CD-2-te & 0.663 & 0.848 & 0.902 & 0.646 \\
             MVB + finetuned on CD-1-tr & CD-2-te  & 0.759 & 0.891 & 0.938 & 0.727 \\
             MVB + finetuned on CD-1-tr \& CD-2-tr & CD-2-te & 0.831 & 0.941 & 0.985 & 0.830 \\
             \hline
              & & \multicolumn{4}{c}{\textbf{G$\rightarrow$Q}} \\
             \hline
             Training & Test & R@1 & R@5 & R@10 & mAP\\
             \hline
             MVB & CD-1-te & 0.656 & 0.843 & 0.907 & 0.558 \\
             MVB + finetuned on CD-1-tr & CD-1-te & 0.826 & 0.924 & 0.963 & 0.757 \\
             MVB & CD-2-te & 0.707 & 0.929 & 0.962 & 0.621 \\
             MVB + finetuned on CD-1-tr & CD-2-te & 0.801 & 0.959 & 0.982 & 0.712 \\
             MVB + finetuned on CD-1-tr \& CD-2-tr & CD-2-te & 0.819 & 0.965 & 0.987 & 0.866 \\
             \hline
        \end{tabular}
    }
    \\ [1ex]
    %*: fine-tuning on train splits
    \caption{Dataset analysis results}
    \label{tab:domain_analysis}
\end{table}

We then extend this analysis to CD-2 to evaluate cross-dataset generalization. A model pretrained on MVB and tested directly on CD-2-te again yields relatively low performance, highlighting the difficulty of generalizing across environments. Fine-tuning on CD-1-tr before testing on CD-2-te improves the results, demonstrating partial transferability between datasets with similar acquisition setups. Finally, further fine-tuning on CD-2-tr leads to the best performance, confirming that dataset-specific adaptation remains necessary even when the acquisition conditions are similar.
Overall, these experiments highlight the presence of a domain gap not only between large-scale external datasets and real-world surveillance data, but also between datasets collected in different environments, despite looking very similar (cf. Fig.~\ref{fig:datasets}). This emphasizes the importance of incremental adaptation when deploying ReID systems in new settings.

\section{Conclusion and Future Work}

%In this paper, we presented a practical multi-camera pipeline for luggage ReID in realistic surveillance environments. 
%Rather than proposing a novel ReID architecture, 
This work focused on studying how detection, tracking, feature selection and aggregation strategies can be combined and adapted in a pipeline for multi-camera luggage ReID in realistic
real-world scenarios and environments.
Through extensive experiments and ablation studies, we analyzed the impact of camera configurations, temporal aggregation, shot selection and dataset adaptation. The results highlight the importance of multi-camera observations and feature aggregation for improving robustness against occlusions and missed detections.
Future work will investigate unsupervised domain adaptation to maintain performance while deploying in new environments without any additional annotation cost.
%Future work will investigate more robust aggregation strategies, cross-camera track association in unconstrained scenes, and larger-scale evaluations in more diverse surveillance environments.

\section*{Acknowledgment}

This work was made possible by the use of FactoryIA Supercomputer (funded by the Ile-de-France Regional Council) and partly funded by the European Union (Project 101121309—BAG-INTEL). Views and opinions expressed are, however, those of the author(s) only and do not necessarily reflect those of the European Union or the European Research Council Executive Agency. Neither the European Union nor the granting authority can be held responsible for them.

%This work was partly funded by the European Union (Project 101121309—BAG-INTEL). Views and opinions expressed are, however, those of the author(s) only and do not necessarily reflect those of the European Union or the European Research Council Executive Agency. Neither the European Union nor the granting authority can be held responsible for them. This work was made possible by the use of FactoryIA Supercomputer (funded by the Ile-de-France Regional Council).

\bibliographystyle{IEEEtran}
\bibliography{references}

\end{document}